\documentclass[a4paper,twoside]{article}

\usepackage{epsfig}
\usepackage{subcaption}
\usepackage{calc}
\usepackage{amssymb}
\usepackage{amstext}
\usepackage{amsmath}
\usepackage{amsthm}
\usepackage{multicol}
\usepackage{pslatex}
\usepackage{apalike}
\usepackage{comment}
\usepackage{algorithm2e}
\usepackage[bottom]{footmisc}

\usepackage{newtxtext}
\usepackage{newtxmath}
\usepackage{array}
\usepackage{hyperref}
\usepackage[colorinlistoftodos]{todonotes}

\newcommand{\imgwidth}{0.42\linewidth}

\usepackage{SCITEPRESS}     

\begin{document}

\title{A Simple Gripper Interface for Simulator-Agnostic Cloth Manipulation}


\author{
\authorname{
Abhilash Nayak\sup{1,*}\orcidAuthor{0000-0001-6228-203X},
Franco Coltraro\sup{1}\orcidAuthor{0000-0002-9149-950X},
Maria Alberich-Carrami\~nana\sup{2}\orcidAuthor{0000-0003-2749-4875}
and Carme Torras\sup{1}\orcidAuthor{0000-0002-2933-398X}
}
\affiliation{
\sup{1}Institut de Rob\`otica i Inform\`atica Industrial, CSIC-UPC, Barcelona, Spain\\
\sup{2}Departament de Matem\`atiques, Universitat Polit\`ecnica de Catalunya, Barcelona, Spain
}
\email{
\{anayak, fcoltraro, torras\}@iri.upc.edu, maria.alberich@upc.edu
}
}

\abstract{
This paper presents a grasping model for cloth manipulation specifically tailored to ease the deployment of robotic control methods. The model is robust, fast and easy to implement avoiding at the same time contact and friction considerations between the gripper and the cloth in favor of simple positional constraints. The gripper is described by its pose, jaw state, and an attached grasping volume. Two kinds of grasping volumes are considered: an axis-aligned box to simulate a pinch grasping and a square pyramidal volume to simulate point grasping. When the gripper closes, the discrete cloth positions lying inside this volume are selected, stored in the local gripper frame, and then transported with the gripper motion. A simple squeezing step is also included to progressively move the selected cloth positions toward the center of the grasping region, avoiding an instantaneous displacement at closure. The model can be used in any simulator as it only requires access to discrete cloth positions and a mechanism for imposing target positions as constraints. We implement our grasping model in conjunction with a constraint-based inextensible cloth simulator, where grasping is implemented as moving positional equality constraints coupled with stretch, shear, collision, and table contact projection steps. The same gripper trajectory is applied on a robot arm to fold a real piece of cloth, serving as a simple bridge between simulation and physical cloth manipulation and showcasing the realism and practicality of our idealized grasping model.}


\onecolumn \maketitle \normalsize \setcounter{footnote}{0} \vfill

\section{Introduction}
\label{sec:introduction}
Robotic cloth manipulation is one of the major challenges in the field of robotics because textiles are highly deformable, their configuration space is very high-dimensional, and they are greatly affected by even the slightest contact. Unlike rigid objects, a piece of cloth or garment cannot be described only by a pose. Its shape changes continuously and small changes in material parameters, grasping, or trajectory execution may produce different final configurations. This makes cloth manipulation difficult for planning, control, learning, and sim-to-real transfer~\cite{yin2021deformable,longhini2024unfolding}.

Robotic cloth manipulation generally incorporates a gripper to interact with the textile. A gripper controls the cloth only through a small set of contact regions. Therefore, the choice of grasp point, the size of the grasp support, and the orientation of the gripper strongly influence the subsequent cloth motion. A realistic treatment of this interaction would require the model to handle contact geometry, pressure distribution, friction, local deformation, slip, and tactile feedback~\cite{yin2021deformable,borras2020grasping}. Such contact-rich models are useful for high-precision simulations, but they are often computationally expensive, simulator-dependent and not feasible for (near) real-time control~\cite{li2020ipc,li2021cipc,coltraro2024collision}.

In this paper, we introduce a simple kinematic gripper model for cloth manipulation. The gripper only has two time-varying attributes: pose and jaw state. Two different kinds of grasping volumes are used to emulate pinch and point grasp. Since the model only requires access to discrete cloth positions, the gripper can be coupled with any constraint based cloth simulator. For instance, the gripper can be used to grasp particles in particle-based simulators like Flex~\cite{nvidiaflex} or SoftGym~\cite{lin2020softgym}, and shell vertices in IPC (Incremental Potential Contact) and C-IPC (Codimensional IPC) based simulators~\cite{li2020ipc,li2021cipc}.  

The motivation is not to replace a precise contact-rich simulation. Instead, the goal is to provide a lightweight interface between a robot gripper and a discrete cloth model which can then be easily used to perform control. 

From a hybrid-intelligence perspective, the proposed interface is only a physics-based model. Nevertheless, its low-dimensional action allows an explicit representation, while the mismatch caused by uncertain
cloth parameters, unmodeled contact, or sim-to-real effects can be assigned to a data-driven residual model.

In this work, the gripper is used with a constraint based inextensible cloth simulator~\cite{coltraro2022inextensible,coltraro2024collision} for folding experiments. Different ways of grasping a corner of a textile to fold it are compared based on the grasp support and orientation of the gripper showcasing their critical importance. Finally, a robot arm is used to perform the same action to show the simplicity and realism of using the model in real life applications.

\subsection{Contributions}
\begin{itemize}
    \item  We introduce a simulator-agnostic kinematic gripper model described only by a pose, a jaw state, and an attached grasping volume. The selected discrete cloth positions are stored in the local gripper frame, progressively squeezed during closure, and transported with the gripper motion for easy use in control applications.
    \item We show how the gripper can be coupled very efficiently to any constraint-based cloth simulator as moving positional equality constraints.
    \item We evaluate eight cases of folding a cloth by its corner with different grasps and gripper orientations and compare their final folding error. 
    \item We demonstrate how a simulation-designed gripper trajectory can be easily executed by a Franka robot arm, providing a simple bridge between simulation and real cloth manipulation.
\end{itemize}

\section{Related works}
\label{subsec:related_work}

Robotic cloth manipulation is strongly affected by how the textile is grasped. Borr{\`a}s et al.~\cite{borras2020grasping} provide a grasping-centered analysis of how local constraints deform cloth, while geometric and trajectory-based folding methods commonly abstract the interaction through selected cloth points or small regions \cite{li2015folding}. More recent work has increasingly incorporated
learning and model-based control for dynamic cloth manipulation \cite{caldarelli2026dynamic,longhini2024unfolding}. Caldarelli et al.~\cite{caldarelli2026dynamic} combine a physics-based cloth simulator with a learned Koopman representation and model predictive control. These approaches motivate a gripper interface that remains simple yet physically meaningful across simulation,
learning, and real-robot execution. However, these methods model the gripper-cloth interaction by prescribing the motion of one cloth point or a small set of adjacent points.

Furthermore, manipulation of deformable objects is prone to error since the material properties, friction, and contact parameters are difficult to identify precisely. These effects can be accounted for by combining the physics-based model with data-driven residual components.
Seyyedi et al.~\cite{seyyedi2023mlphysics} classify physics-ML integration into physics-guided ML, ML-guided physics, and mutually guided approaches. Within these categories, deformable-object methods include residual correction and parameter estimation. Liang et al.~\cite{liang2024realtosim}, for example, use learned residual mappings to adapt a physics simulator, while Jiang et al.~\cite{jiang2025phystwin} infer physical properties from observations for physics-based simulation.

The present work focuses exclusively on the white-box component of such a potential hybrid formulation. Specifically, we develop and experimentally evaluate a simulator-agnostic, physics-based gripper
interface in which grasp geometry,  orientation, and motion remain
explicit and physically interpretable. The goal is to retain the essential effect of grasping on the cloth motion while avoiding the cost and models that rely on detailed contact, friction, and finger-cloth interaction. The resulting model is sufficiently realistic to distinguish between different grasp supports and gripper orientations, but remains lightweight and efficient enough for control applications. It therefore provides a practical bridge between idealized point grasp abstractions and more detailed gripper-cloth contact models.

\section{Gripper}
\subsection{Gripper Pose and Jaw State}
The attributes of the gripper are its pose, and its jaw state. 
The gripper pose is represented by a position
$p_g(t)\in\mathbb{R}^3$ and a unit quaternion $q_g(t)$,
\begin{equation}
    T_g(t)=\big(p_g(t),q_g(t)\big).
\end{equation}
Quaternions are used as they provide a compact representation for pose streaming from a robot or simulation interface and are easy to normalize numerically.



To determine whether a cloth node $p_i(t)$ lies within the grasping region, it is first transformed into the gripper coordinate frame using quaternion rotation, yielding $p_i^g(t)$. 

The other time-varying attribute is its discrete jaw state:
\begin{equation}
    s_g(t) \in \{ 0, 1 \}.
\end{equation}
where $s_g = 0$ denotes open jaws and $s_g = 1$ denotes closed jaws. 
\subsection{Grasping region}
The grasping region is a volume rigidly attached to the gripper frame. Node selection is performed in the gripper frame, not in the world frame. Thus, the volume follows the gripper pose: translating or rotating the gripper changes the location and orientation of the grasping region in the world frame, while the local membership test remains unchanged.




\begin{figure}[!t]
  \centering
    \includegraphics[width=0.8\linewidth]{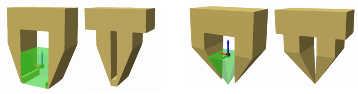}
  \caption{Box-shaped pinch gripper and square-pyramidal point gripper: jaws open and closed.}
  \label{fig:grippers}
 \end{figure}
To grasp a node, two kinds of grippers are used. The first one simulates a pinch grasp with an axis-aligned box shaped grasping region:
\begin{equation}\label{eq:box}
    \mathcal{B}_g(b) = \left\{ p \in \mathbb{R}^3 :|p_r| \leq \frac{b_r}{2}, \quad r \in \{x,y,z\} \right\}
\end{equation}
where $[b_x, b_y, b_z]^T$ are the dimensions of the box, and the gripper coordinate frame is located at the center of the box as shown in Fig.~\ref{fig:grippers}. 
The other gripper simulates a point grasp with a square-pyramid shaped grasping region:
\begin{equation}\label{eq:pyramid}
\mathcal{B}_g(s,h)
=
\left\{
p\in\mathbb{R}^3 :
\begin{array}{l}
0\leq p_z \leq h,\\[2pt]
|p_x| \leq \dfrac{s}{2}\left(1-\dfrac{p_z}{h}\right),\\[2pt]
|p_y| \leq \dfrac{s}{2}\left(1-\dfrac{p_z}{h}\right)
\end{array}
\right\}
\end{equation}
where $s$ is the base side length, $h$ is the height, and the gripper coordinate frame is located at the center of the square base with its $z$-axis passing through the apex, as shown in Fig.~\ref{fig:grippers}. 
To compare the two grasping volumes, the height of the square pyramid is chosen to be equal to half of the box side length along the $z$-axis: $h=b_z/2$.

\subsection{Node Selection}
\begin{figure}[!t]
  \centering
    \includegraphics[width=0.8\linewidth]{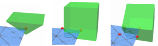}
  \caption{Example of a gripper-attached grasping volumes overlapping a quadrilateral cloth mesh. Nodes directly inside the volume are selected first. If an edge center or a face center lies inside the volume, the corresponding edge nodes or face nodes are also added to the grasp support. This avoids missing a grasp on coarse meshes when the volume intersects an element but does not contain a node.}
  \label{fig:grasp_support}
 \end{figure}
A node $i$ is selected if it lies in the grasp region:
\begin{equation}
    p_i^g(t) \in \mathcal{B}_g.
\end{equation}
Here $\mathcal{B}_g$ denotes the chosen grasping region, either the box $\mathcal{B}_g(b)$ or the pyramid $\mathcal{B}_g(s,h)$.
Selecting only nodes this way can be unreliable for coarse meshes. A small grasp box may intersect a face or an edge without containing any node. Therefore, the implementation also checks face centers and edge centers.



For a quadrilateral face $f=(i_1,i_2,i_3,i_4)$ we use
$p_f^g=\frac{1}{4}\sum_{\ell=1}^{4}p_{i_\ell}^g$, while for an
edge $e=(i,j)$ we use $p_e^g=\frac{1}{2}(p_i^g+p_j^g)$.
If the corresponding center lies inside $\mathcal{B}_g$, the
incident face or edge nodes are added to the grasp support.

The final grasp support is therefore
\begin{equation}\label{eq:support_union}
\mathcal{I}_g = \mathcal{I}_g^n \cup \mathcal{I}_g^f \cup \mathcal{I}_g^e,
\end{equation}
where $\mathcal{I}_g^n$ comes from node tests, $\mathcal{I}_g^f$ from face-center tests, and $\mathcal{I}_g^e$ from edge-center tests. Figure~\ref{fig:grasp_support} illustrates the selection rule on a meshed cloth.

The grasp support is detected only during the jaw transition from open~($s_g(t) = 0$) to closed~($s_g(t) = 1$). Let $t_c$ be the closing time. At $t_c$, the set $\mathcal{I}_g$ is computed using~\eqref{eq:support_union}. For each selected node $i \in \mathcal{I}_g$, its position in gripper coordinate frame $p_i^g$ is stored. These nodes remain controlled until the gripper opens. When it does, the grasp support is cleared. The cloth then resumes free motion under its internal dynamics, collision and gravity. 

\subsection{Squeezing model}
In reality, closing the jaws does not only attach the cloth to the gripper. It also compresses the captured textile in the jaw-closing direction. A purely instantaneous attachment may produce a sudden jump in the cloth configuration, especially when the detected nodes are not exactly centered between the jaws. To avoid this, we implement a simple squeezing model. 

Let $e_x$ denote the local jaw-closing direction. For each grasped node $i \in \mathcal{I}_g$, the corresponding position vector $p_i^0$ is squeezed towards the center $c$ of the grasp region. Since they are expressed in the gripper coordinate frame, $c$ is simply its origin. Thus, the squeezed local target is defined as
\begin{equation}\label{eq:squeeze_target}
p_i^\star = p_i^0 + \sigma\left(c_x -p^0_{i,x}\right)e_x + \varepsilon e_z.
\end{equation}
Here $\sigma \in [0,1]$ is the squeeze amount and $\varepsilon > 0$ is a small lift in the local $z$-direction. Since $a_i^0$ is expressed in the gripper frame, $c_x$ must also be interpreted in the gripper frame. In particular, $c_x=0$ only when the grasping region is centered at the gripper-frame origin. If the grasping region is offset, for example near the fingertip, then $c_x$ is the corresponding local offset.

Let $\alpha(t)\in[0,1]$ be a scalar interpolation variable. The local target imposed by the gripper is
\begin{equation}\label{eq:squeeze_interpolation}
p_i(t) = (1-\alpha(t))p_i^0 + \alpha(t)p_i^\star.
\end{equation}
After closure, $\alpha$ is increased over several simulation steps until it reaches $1$. This produces a progressive squeeze rather than an abrupt displacement.
\begin{figure}[!t]
  \centering
    \includegraphics[width=0.5\linewidth]{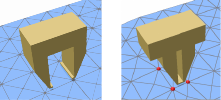}
  \caption{Comparison between grasping before and after squeezing using the proposed gradual squeezing model with $\sigma = 0.7$. Without squeezing, controlled nodes would have stayed where they are, which can create a visually abrupt attachment with the gripper. With squeezing, selected nodes are progressively moved toward the jaw-closing direction, producing a realistic jaw closure.}
  \label{fig:squeeze}
 \end{figure}
Figure~\ref{fig:squeeze} shows the state of the cloth before and after squeezing. It is noteworthy that this is still not a full contact model. It does not compute pressure, friction, or local material deformation. It is only a kinematic approximation of the compression induced by jaw closure.

\subsection{Grasping as a Constraint}
At each time step after the gripper has closed, the desired world position of every grasped node is computed from the gripper position $p_g(t)$ and quaternion $q_g(t)$ as
\begin{equation}\label{eq:desired_world}
u_i(t) = p_g(t) + \operatorname{vec}\left(q_g(t)\otimes \widehat{p_i^g(t)}\otimes q_g^{*}(t)\right), \, \, i\in\mathcal{I}_g.
\end{equation}
Here $\widehat{p_i^g(t)}$ is the pure-quaternion representation of $p_i^g(t)$, $q_g^{*}(t)$ is the quaternion conjugate, $\otimes$ denotes quaternion multiplication, and $\operatorname{vec}(\cdot)$ extracts the vector part.

The grasp constraint is therefore
\begin{equation}\label{eq:node_constraint}
p_i(t)-u_i(t)=0, \qquad i\in\mathcal{I}_g.
\end{equation}
Thus, the gripper imposes a set of moving positional equality constraints on the selected cloth nodes. These constraints can be added to any cloth-simulation system. A position-based simulator may directly project the selected nodes to $u_i(t)$. A force-based simulator may apply spring-damper forces toward $u_i(t)$. A constraint-based simulator may append Eq.~\eqref{eq:node_constraint} to the solver as additional equality residuals.

In this paper, we use a constraint based inextensible cloth simulator~\cite{coltraro2022inextensible,coltraro2024collision}, where the cloth is represented as a quadrilateral mesh. At each time step, the simulator computes an unconstrained inertial update and then projects the cloth configuration to the admissible set defined by cloth and contact constraints. Grasping is added to this formulation as an additional equality constraint given by \eqref{eq:node_constraint}. Thus, the gripper does not directly apply contact forces. Instead, the grasped nodes are treated as controlled points on the cloth, whose positions are prescribed by the gripper pose. This makes the gripper simple while still coupling the grasped nodes to the constrained cloth dynamics. 

The proposed grasping interface introduces only a small computational overhead with respect to the underlying cloth simulator. The grasp support $\mathcal{I}_g$ is computed only once, during the transition
from open to closed jaws. During subsequent motion, each grasped node requires only a rigid transformation from the stored gripper-frame coordinates and the enforcement of one positional equality constraint. Consequently, the additional per-step cost scales linearly with the number of grasped nodes, i.e., $\mathcal{O}(|\mathcal{I}_g|)$. These results are reported in Sec.~\ref{subsec:experiment_simualation}.


\subsection{Gray-Box Extension for Hybrid Intelligence}
\label{subsec:hybrid_extension}
The gripper model is a prior to be used with a learned residual model. We could learn the error in the prediction of the physics model or corrections to the simulator parameters. If $x_t$ denotes the cloth state, $u_t=(p_g(t),q_g(t),s_g(t))$ the gripper command, and $\theta$ the simulator parameters, the physics-based prediction done in a simulator can be written abstractly as
\begin{equation}
    \widetilde{x}_{t+1}
    =
    F_{\mathrm{phys}}(x_t,u_t;\theta).
\end{equation}
A gray-box extension can retain this prediction and learn only the residual mismatch:
\begin{equation}
    \widehat{x}_{t+1}
    =
    \widetilde{x}_{t+1}
    +
    r_{\phi}(z_t,u_t,\widetilde{x}_{t+1}),
\end{equation}
where $z_t$ denotes available observations and $r_{\phi}$ is a data-driven residual model. 
An alternative is parameter correction,
\begin{equation}
    \theta_t = \theta_0 + \Delta_{\phi}(z_{0:t}).
\end{equation}
Both formulations retain the proposed gripper interface and account for unmodeled deformation, contact, slip, or
sim-to-real effects to data
\cite{seyyedi2023mlphysics,liang2024realtosim,jiang2025phystwin}.

The present experiments evaluate the physical model. Training and benchmarking a learned residual model are left for future work.

\section{Cloth Folding Experiment}
\subsection{Comparison of Different Kinds of Grasping} \label{subsec:experiment_simualation}
\begin{figure}[!t]
  \centering
    \includegraphics[width=0.8\linewidth]{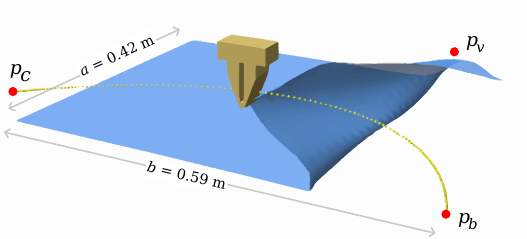}
  \caption{Parabolic trajectory, shown in yellow, defined by a quadratic B\'{e}zier curve with three control points, shown in red.}
  \label{fig:fold}
 \end{figure}

The goal of this experiment is to simply fold a flat cloth in two, thereby evaluating different grasping strategies on the final configuration of the folded cloth. All cases use the same cloth, the same simulator parameters, and the same pulling trajectory. The only differences between the cases are the grasping volume, the number and arrangement of selected cloth nodes, and whether the gripper keeps or changes its orientation during the fold.

The cloth is modeled as a rectangular textile of side lengths $a=0.42$ m and $b=0.59$ m, as shown in Fig~\ref{fig:fold}. The quadrilateral mesh is generated with a resolution of $n_a=23$ and $n_b=28$. 

The gripper starts at the selected corner and follows the trajectory defined by a quadratic B{\'e}zier curve in the $y$-$z$ plane as shown in Fig.~\ref{fig:fold}. The control points are grasp position $p_b$, final gripper position $p_c$, and an intermediate point $p_v$. 
The gripper position during the pulling phase is then given by
\begin{equation}
	p_g(s)=(1-s)^2p_b+2(1-s)sp_v+s^2p_c.
\end{equation}
In the implementation, the $x$-coordinate remains fixed at the grasped corner value, while the parabolic motion is applied to the $y$ and $z$ coordinates. For the rotated cases, the gripper orientation is interpolated between the initial grasp orientation and the final rotated orientation.

\begin{table*}[!t]
\caption{Comparison of the eight corner-grasping cases. \href{https://drive.google.com/file/d/1ggT3ZJ_SUYwwVa9t1cCsQMNDhejui05m/view?usp=sharing}{Link} to video.}\label{tab:corner_grasping_cases} \centering
\begin{tabular}{
|>{\centering\arraybackslash}m{0.8cm}
|>{\centering\arraybackslash}m{1.2cm}
|>{\centering\arraybackslash}m{1.5cm}
|>{\centering\arraybackslash}m{1.3cm}
|>{\centering\arraybackslash}m{1.7cm}
|>{\centering\arraybackslash}m{5.1cm}|}
  \hline
  Case & Grasp support & Orientation change & Grasping volume & Error & Configuration (grasped nodes in red) \\
  \hline
  C1 & 1 node & No & Pyramid & 0.02230371  & \includegraphics[width=\imgwidth]{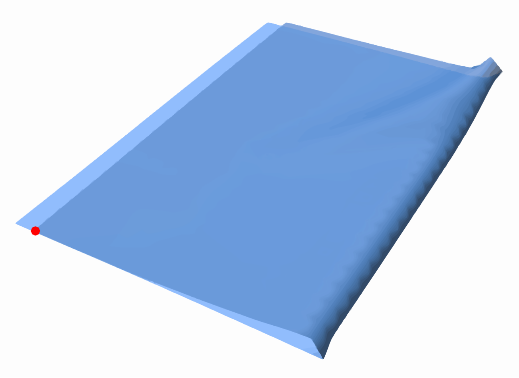} \\
  \hline
  C2 & 2 nodes & No & Box & 0.02269459 & \includegraphics[width=\imgwidth]{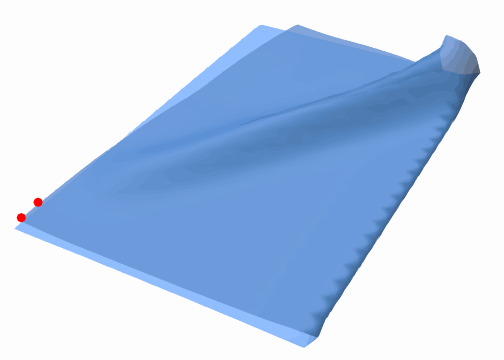} \\
  \hline
  C3 & 2 nodes & No & Box & 0.08618127 & \includegraphics[width=\imgwidth]{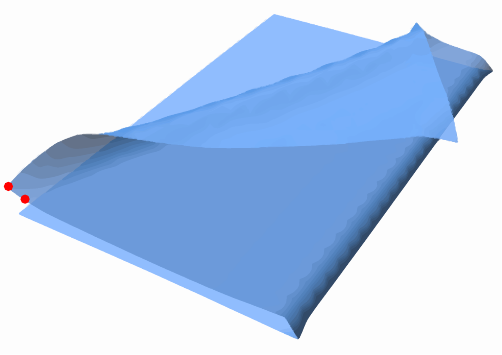} \\
  \hline
  C4 & 4 nodes & No & Box & 0.09149399 & \includegraphics[width=\imgwidth]{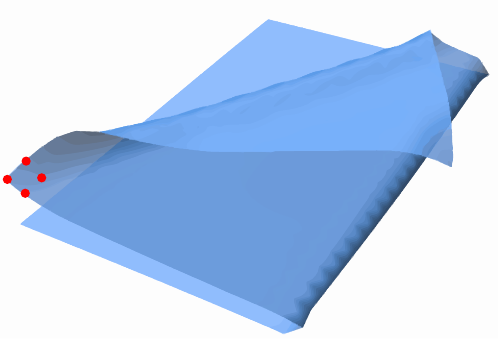} \\
  \hline
  C5 & 1 node & Yes & Box & 0.02994773 & \includegraphics[width=\imgwidth]{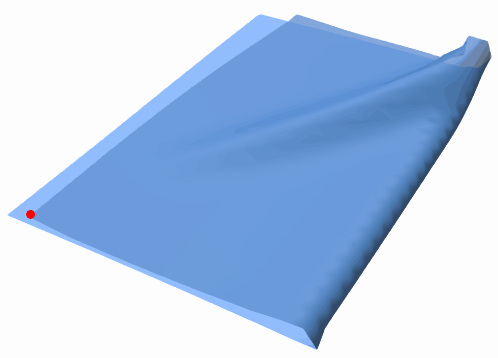} \\
  \hline
  C6 & 2 nodes & Yes & Box & 0.02412322 & \includegraphics[width=\imgwidth]{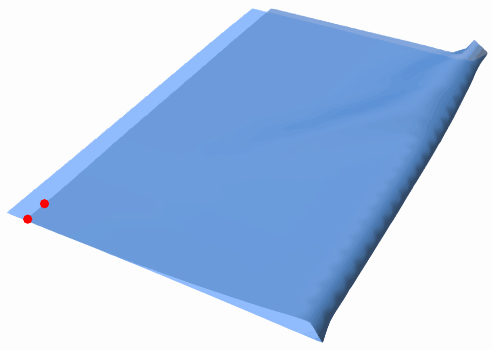} \\
  \hline
  C7 & 2 nodes & Yes & Box & 0.00881858 & \includegraphics[width=\imgwidth]{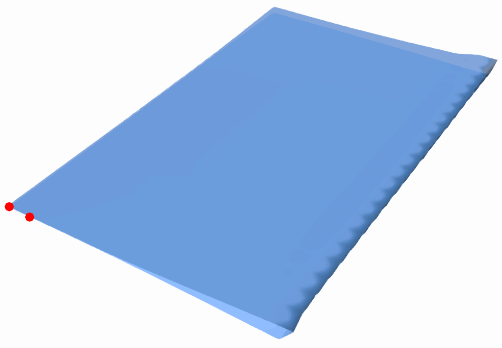} \\
  \hline
  C8 & 4 nodes & Yes & Box & 0 & \includegraphics[width=\imgwidth]{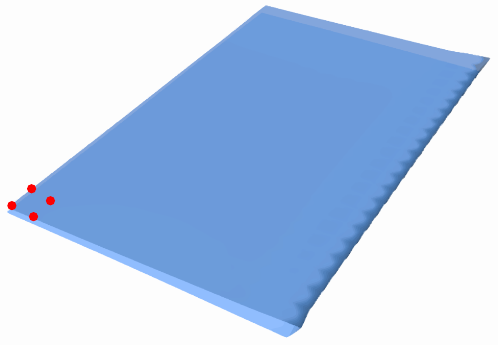} \\
  \hline
\end{tabular}
\end{table*}

Initially, the folding error was defined with respect to an analytically constructed ideal fold. However, in the present set of experiments, Case C8 in Table~\ref{tab:corner_grasping_cases} produces a final configuration that visually matches the intended fold most closely. We therefore use the final configuration of C8 as the reference fold to measure how far each grasping strategy deviates from the best folded configuration obtained in the experiment.

Let $X(t)=\{p_1(t),\ldots,p_n(t)\}$ denote the simulated cloth configuration, $p_i(t)\in\mathbb{R}^3$ is the position of the $i$-th node of the quadrilateral cloth mesh at time $t$. Let $X^\star=\{p_1^\star,\ldots,p_n^\star\}$ denote the ideal final fold configuration. The folding error is measured as the normalized root-mean-square distance to the ideal folded cloth,
\begin{equation}\label{eq:fold_error}
E(t)=\frac{1}{d}\sqrt{\frac{1}{n}\sum_{i=1}^{n}\|p_i(t)-p_i^\star\|^2}, \, \, d=\sqrt{a^2+b^2}.
\end{equation}
A lower value of $E(t)$ indicates that the simulated cloth is closer to the ideal fold. By construction, Case C8 is the reference case and therefore has zero final error under this metric.

We compare eight cases as shown in Table~\ref{tab:corner_grasping_cases}. The evolution of the cloth for all these cases are shown in the \href{https://drive.google.com/file/d/1ggT3ZJ_SUYwwVa9t1cCsQMNDhejui05m/view?usp=sharing}{video}. The final error for each simulation links each case to the corresponding configuration of the folded cloth. We consider a single-node grasp in C1 and C5, a two-node grasp in C2, C3, C6, C7, and a four-node grasp in C4 and C8, where all vertices of the corner quadrilateral are selected. C1-C4 are tested without rotating the gripper and C5-C8 are tested by rotating the gripper similar to how a human would ideally fold the cloth. Cases C1 and C2 use the same grasping pose but different grasping volumes: C1 uses the square pyramid and grasps only one node while C2 grasps two nodes because of the box-shaped grasping region. This shows that, for the same grasp position $p_b$, different grasping volumes lead to different grasp supports.

\begin{figure}[!t]
    \centering
    \includegraphics[width=\linewidth]{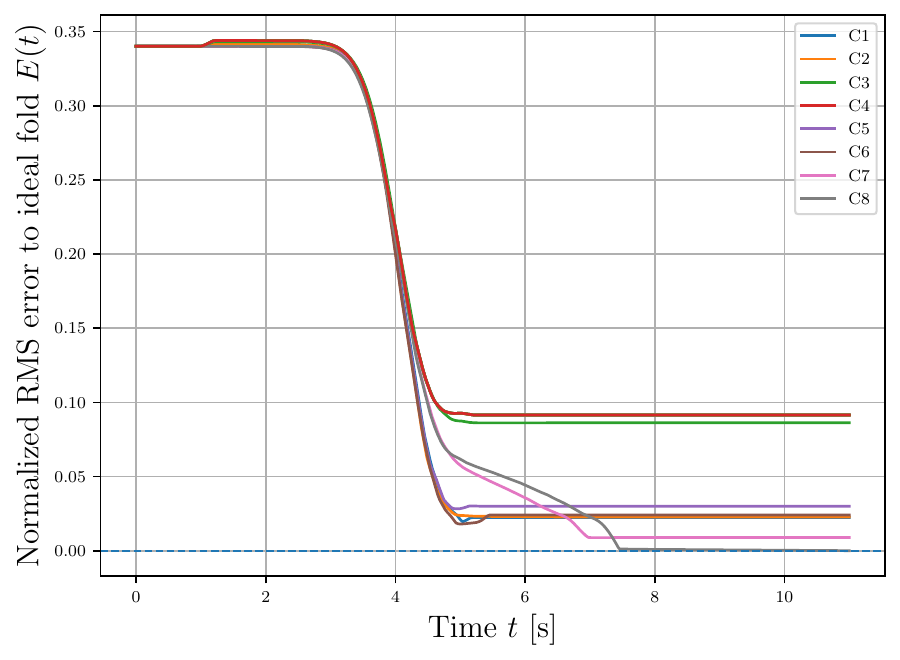}
    \caption{Normalized RMS error $E(t)$ for the eight corner-grasping cases. All simulations use the same $p_{\mathrm{corner}}$ and the same parabolic pulling trajectory. The curves differ only in the grasp support, gripper orientation, and grasping volume.}
    \label{fig:error_curves}
\end{figure}

Figure~\ref{fig:error_curves} shows the full time evolution of $E(t)$ for all cases. Cases C5-C7 produce the lowest errors, indicating that, for this particular parabolic motion, rotating the gripper during the fold can lead to a better fold. This is expected because rotation does not only translate the grasped cloth region, but also imposes a local orientation on the selected patch. Among these cases, the ideal case C8 can be explained by its larger and more coherent grasp support: the four selected corner nodes constrain a small cloth patch rather than only a point or an edge. As a result, the gripper motion is transferred more consistently to the cloth, reducing local twisting near the corner and producing a final configuration closer to the reference fold.
The measured computational overhead increased approximately linearly with the number of grasped nodes. For $|\mathcal{I}_g|=\{1,2,4\}$, the gripper-interface computation required
0.17, 0.20, and 0.22~ms per simulation step, respectively, corresponding to
0.5 - 1.5\% of the total simulation time.

\subsection{Cloth Folding Using a Robot Arm}
\label{subsec:robot_folding}

\begin{figure*}[!t]
    \centering
    \includegraphics[width=0.7\textwidth]{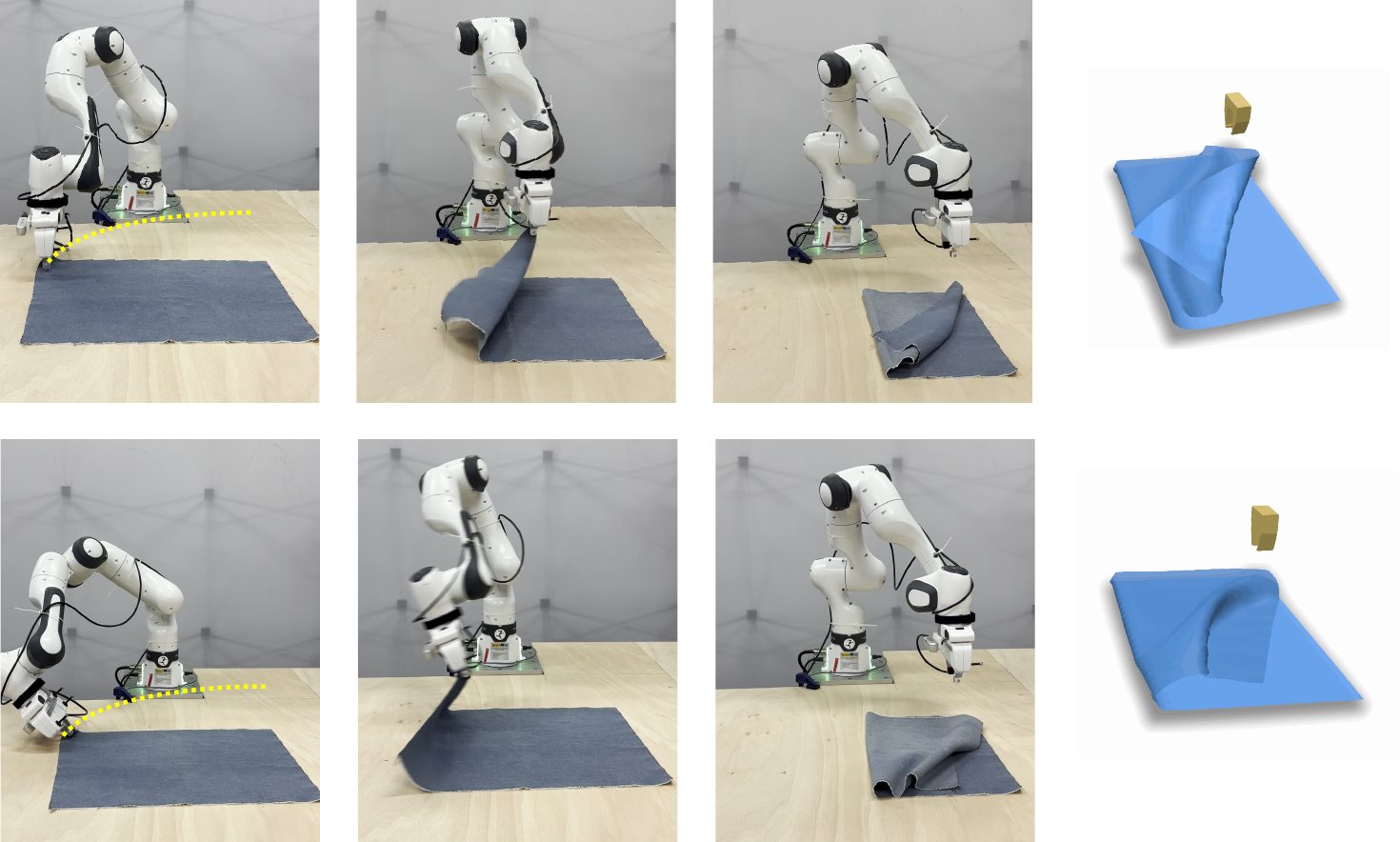}
    \caption{
        Selected frames from the cloth-folding experiment performed with the Franka robot. The top row shows the execution when the gripper orientation is kept constant (\href{https://drive.google.com/file/d/12Og5HotJZxP6ZBztexCDV2uoWhlleT6a/view?usp=drive_link}{Link} to video). The bottom row shows
        the corresponding execution when the gripper orientation is changing along the trajectory (\href{https://drive.google.com/file/d/1_xCFmVMwiFhRtI9xfJg5TJKK2JW4T7dC/view?usp=sharing}{Link} to video). The last image shows the final configuration of the cloth in simulation.
    }
\label{fig:franka_frames}
\end{figure*}

The previous section used simulation to compare different grasping strategies under controlled conditions. The experiment in this section uses a similar folding motion with an objective to show that the proposed gripper representation can be transferred naturally from simulation to a real robot arm. 

For this experiment, we used a laboratory cloth and performed a partial system identification to approximate its simulator parameters. The estimated parameters (refer~\cite{coltraro2022inextensible}) include the cloth density $\rho=0.3046~\mathrm{kg\,m^{-2}}$, virtual gravitational mass $\delta=0.139~\mathrm{kg\,m^{-2}}$, and damping coefficient $\alpha=0.416~\mathrm{s^{-1}}$. The bending stiffness $\kappa=4\times10^{-5}$ and the friction coefficients $\mu_f=0.4$ (floor) and $\mu_s=1.3$ (cloth) were adjusted to obtain qualitatively similar interaction with the table on which the cloth is placed. 

Consequently, the physical experiment should be interpreted under parameter uncertainty rather than as a fully calibrated simulator.

Uncertainty in damping, bending stiffness and friction affects dissipation, fold curvature and contact behavior,
respectively. These discrepancies constitute natural sources of sim-to-real residual error and motivate the residual or
parameter correction formulations in
Sec.~\ref{subsec:hybrid_extension}.


After this partial identification, the same corner-folding motion was executed in simulation and on a Franka robot arm equipped with a parallel-jaw gripper. In the robot implementation, the end-effector trajectory is executed using MoveIt Cartesian controller in ROS~2, while the gripper jaw state is sent as a separate open/close command.

Figure~\ref{fig:franka_frames} shows representative frames from the robot experiment and the corresponding simulation. The current results are qualitative, but they show that the same gripper-level command can be used in both settings. This experiment therefore serves as a proof of transfer of the proposed gripper interface. It shows that the interface is not tied to the simulator, while also showing the role of gripper orientation, which is often ignored in literature~\cite{borras2020grasping,li2015folding,caldarelli2026dynamic}. The \href{https://drive.google.com/file/d/12Og5HotJZxP6ZBztexCDV2uoWhlleT6a/view?usp=drive_link}{constant-orientation} and \href{https://drive.google.com/file/d/1_xCFmVMwiFhRtI9xfJg5TJKK2JW4T7dC/view?usp=sharing}{changing-orientation} videos show the evolution of cloth in simulation, further asserting the importance of gripper orientation.


\section{Conclusions and Future Work}
This paper introduced a simple kinematic gripper interface for cloth manipulation. The gripper is defined by a pose, a jaw state, and a grasping volume attached to the gripper frame. When the jaws close, the selected discrete cloth positions are stored in local coordinates, progressively squeezed, and later transported with the gripper motion. The formulation is intentionally minimal: it only requires access to discrete cloth positions and a mechanism for imposing target positions or constraints. For this reason, the same interface can be connected in a direct and natural way to particle-based, FEM-based, projection-based, or constraint-based cloth simulators.

The model was demonstrated with a constraint-based inextensible cloth simulator. The corner-folding experiments show that even in a simple folding task, the grasp support and gripper orientation can significantly affect the final cloth configuration. This is important because many cloth manipulation methods use point grasp abstractions or prescribe the motion of selected cloth points while ignoring the orientation and finite support of the grasp. Our results show that these quantities can strongly influence the outcome and should be represented explicitly.

A Franka arm was also used to perform a similar folding motion with a real piece of cloth. Although the proposed grasping model is idealized, the robot experiment shows that it can reproduce realistic folding behavior when combined with an approximate identification of the cloth parameters. This supports the main goal of the model: to be simple and efficient, but still meaningful enough for control and learning applications, where a small number of variables with clear physical interpretation is preferable.

The main limitation of the proposed model is that it is not a contact-rich gripper model. The gripper does not collide with the cloth through its external geometry. Only the cloth positions selected inside the grasping volume at closure are controlled. As a consequence, the model cannot represent pushing with the gripper surface, sliding contact along the gripper body, frictional slip inside the jaws, pressure distribution, finger compliance, or tactile feedback. These simplifications are acceptable for trajectory-level folding experiments, but not for high-precision manipulation tasks where local contact mechanics determine success.

Future work will consider two extensions. The first is to use the proposed gripper as a simulator-agnostic interface for learning and control, including model-based control, reinforcement learning, and residual sim-to-real correction. The second is to add contact-rich gripper interaction for tasks requiring high precision. In that setting, the present kinematic model can serve as a fast baseline, while more detailed contact solvers or tactile feedback can be activated only when local finger and cloth interaction is necessary.


\section*{\uppercase{Acknowledgements}}

A. Nayak is supported by the European Commission's NextGenerationEU initiative through the MomentumCSIC Programme: Develop Your Digital Talent, project MMT24-IRII-01.
M. Alberich-Carramiñana and F. Coltraro are partially supported by the European Projects HORIZON-CL4-2024-DIGITAL-EMERGING-01-101189600 (FlexCycle) and HORIZON-WIDERA-2023-ACCESS-02-01-101159522 (ROMANDIC).
M. Alberich-Carramiñanana is also partially supported by the Spanish State Agency project PID2023-146936NB-I00 funded by MICIU/AEI/10.13039/501100011033, and by the AGAUR project 2021 SGR 00603.

\bibliographystyle{apalike}
{\small
\bibliography{example}}



\end{document}